\documentclass[lettersize,journal]{IEEEtran}
\usepackage{amsmath,amsfonts}
\usepackage{amssymb}
\usepackage{multirow}
\usepackage{algorithmic}
\usepackage{algorithm}
\usepackage{tabularx}
\usepackage{array}
\usepackage[caption=false,font=normalsize,labelfont=sf,textfont=sf]{subfig}
\usepackage{textcomp}
\usepackage{stfloats}
\usepackage{url}
\usepackage{verbatim}
\usepackage{graphicx}

\begin{document}

\title{UnityGraph: Unified Learning of Spatio-temporal features for Multi-person Motion Prediction}

\author{Kehua Qu†, Rui Ding†*, Jin Tang
\thanks{Kehua Qu, and Rui Ding are with Information Engineering College, Capital Normal University, Beijing 100048, China. Email: 2221002103@cnu.cn, 5758@cnu.cn. Jin Tang is with the School of Intelligent Engineering and Automation, Beijing University of Posts and Telecommunications, Beijing 100876, China. E-mail: tangjin@bupt.edu.cn. }
\thanks{*Corresponding authors.}
\thanks{†These authors contributed equally to this work.}}

\markboth{Journal of \LaTeX\ Class Files,~Vol.~14, No.~8, APRIL~2024}%
{Shell \MakeLowercase{\textit{et al.}}: A Sample Article Using IEEEtran.cls for IEEE Journals}


\maketitle

\begin{abstract}
Multi-person motion prediction is a complex and emerging field with significant real-world applications. Current state-of-the-art methods typically adopt dual-path networks to separately modeling spatial features and temporal features. However, the uncertain compatibility of the two networks brings a challenge for spatio-temporal features fusion and violate the spatio-temporal coherence and coupling  of human motions by nature. To address this issue, we propose a novel graph structure, UnityGraph, which treats spatio-temporal features as a whole, enhancing model coherence and coupling. Specifically, UnityGraph is a hypervariate graph based network.
The flexibility of the hypergraph allows us to consider the observed motions as graph nodes. We then leverage hyperedges to bridge these nodes for exploring spatio-temporal features. This perspective considers spatio-temporal dynamics unitedly and reformulates multi-person motion prediction into a problem on a single graph. Leveraging the dynamic message passing  based on this hypergraph, our model dynamically learns from both types of relations to generate targeted messages that reflect the relevance among nodes. Extensive experiments on several datasets demonstrates that our method achieves state-of-the-art performance, confirming its effectiveness and innovative design.
\end{abstract}


\section{Introduction}
\begin{figure*} 
	
	\centering
	
	\includegraphics[width=1\textwidth]{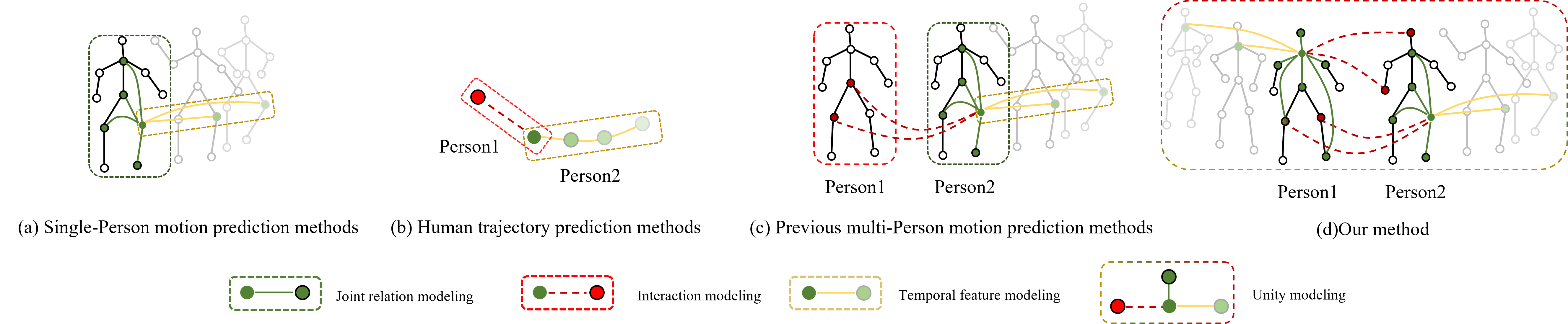}

	\caption{Comparison between our method with single-person prediction methods \cite{zhong2022spatio,dang2021msr}, human trajectory prediction methods \cite{xu2022groupnet,li2020evolvegraph} and traditional multi-person motion prediction methods \cite{wang2021multi,xu2023joint,peng2023trajectory}. (a)Single-person prediction methods focus on modeling joint relations, neglecting interactions within the group. (b)Human trajectory prediction methods lack the representation of 3D pose. (c)Traditional multi-person motion prediction methods employ multiple sub-networks to  capture spatial and temporal features separately. These methods inevitably diminish spatio-temporal coupling and consistency. (d)Our method unifies the learning of spatio-temporal features within a single network for multi-person motion prediction. For clarity, edges that connect nodes across different frames are omitted.}
	\label{intro1}

\end{figure*}

Multi-person motion prediction aims to predict future human motion sequences for multiple individuals based on historical sequences. Single-person motion prediction methods \cite{ nie2023triplet,10064318,saadatnejad2023generic} focus solely on modeling an individual's spatial joints relation or trajectory features in the temporal domain, neglecting the interactions between different individuals, as shown in Fig. \ref{intro1} (a). In contrast, multi-person motion prediction carries more practical significance, as it is more common for multiple people to be present in one scene in the real world. At the same time, multi-person motion prediction is also more challenging because of the complicated interactions between individuals and plays a significant role in many real-world applications, such as autonomous driving \cite{tang2023collaborative,fang2023tbp}, robotics \cite{gao2021human,8460651}, surveillance systems \cite{xu2022remember}.

Most current researches \cite{xu2022groupnet,li2020evolvegraph} involving multi-person scenes focus on trajectory prediction, which model the agent as a single 2D point on the ground plane and exploit the temporal features and the spatial interactions among different person to make prediction, as shown in Fig. \ref{intro1} (b). However, these methods are insufficient for 3D tasks such as motion prediction, which requires detailed body pose information. Adeli et al. \cite{9709907} first combine scene context to model interactions between humans and objects in 3D task. Subsequently, more approaches \cite{wang2021multi,xu2023joint,peng2023trajectory} are being developed to capture spatio-temporal features more effectively. However, previous methods usually employ dual-path networks to learn temporal and spatial features separately, as shown in Fig. \ref{intro1} (c). Given the variations and diverse action types in human motion data, this decoupled modeling strategy inherently disrupts the real world unified spatio-temporal interdependencies, making it challenging to accurately capture the cross-dependencies of spatio-temporal relationships and  potentially limiting the accuracy and realism of the predictions \cite{yi2024fouriergnn, zhong2022spatio}.

\begin{figure*} 
	
	\centering
	
	\includegraphics[width=1\textwidth]{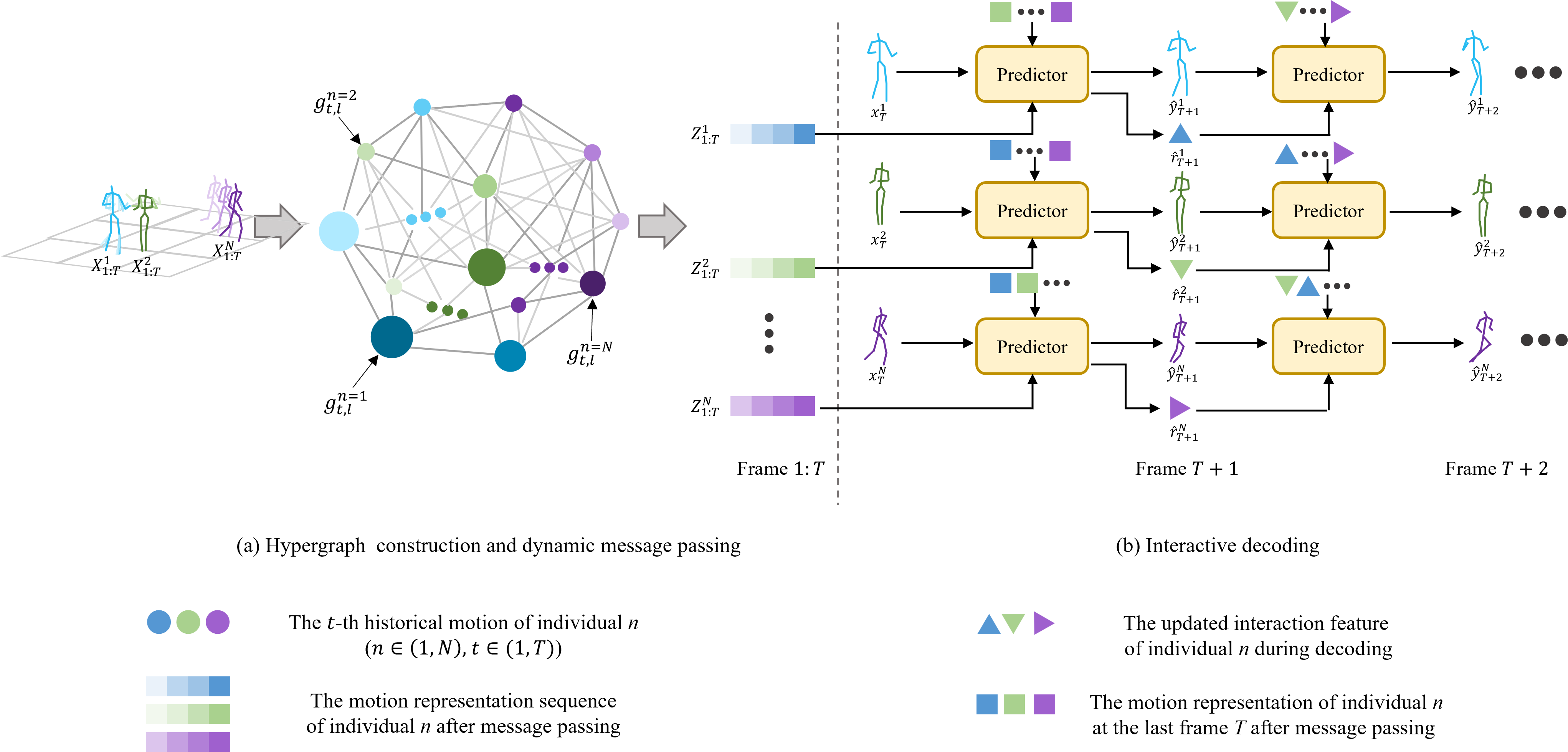}

	\caption{The framework of our network. (i) Each motion representation of individual in obseved frame is considered as a node of the graph. And the various hyperedges denote relation in the temporal or spatial dimensions. Message passing propagates spatio-temporal features through these edges. (ii) The interactive decoding incorporates historical features, such as the last frame of observed motion and the motion sequence after message passing, along with updated relations through reasoning at each frame.}
	\label{arc}

\end{figure*}

To address the above issue, we propose a novel graph structure named UnityGraph based on hypervariate graph to learn human motion features with the spatio-temporal consistent perspective, as shown in Fig. \ref{intro1} (d). The core idea of the hypervariate graph is to construct a space-time fully-connected structure. Specifically, given the observation of scenes involving $N$ persons over a duration of $T$ time steps. The hypergraph can be viewed as a graph with $N \times T$ nodes that are connected by hyperedges, as shown in Fig. \ref{arc} (a). The higher-order connectivity \cite{feng2019hypergraph,10319392} of hyperedge can connect multiple nodes at the same time is utilised to express complex interactions between multiple people over the spatio-temporal domain. Such a special design formulates spatio-temporal features of individuals as node-to-node dependencies in the hypergraph. Different from previous methods that learn spatio-temporal features separately with a dual-path structure, our approach views spatio-temporal correlations as a whole. It abandons the uncertain compatibility of spatial and temporal modeling, constructs adaptive spatio-temporal dependencies. We set three types of hyperedges to learn spatio-temporal features from node-to-node dependencies: i) Short-term hyperedges connect neighboring nodes of the same individual to model short-term dynamics. ii) Long-term hyperedges connect all nodes of the same individual throughout the time series, capturing the long-term dynamics. iii) Spatial hyperedges link different nodes at the same frame to capture interaction features between different individuals. Notably, our strategy of designing hyperedges achieves more accurate predictions than treating the hypergraph as a fully connected graph. This approach ensures spatio-temporal coupling while avoiding the unacceptably large computational effort associated with full connectivity, as discussed in our extensive ablation experiments.

Following the construction of the hypergraph, multi-person motion prediction is reformulated as a prediction task on a single graph. To capture spatio-temporal features from this graph, we introduce dynamic message passing that enhances the spatio-temporal feature transfer between nodes and hyperedges. In the initial node-to-hyperedge phase, nodes transmits information to its associated hyperedges, which update their state by aggregating this information through weighted sums. Subsequently, hyperedges send their aggregated information back to the connected nodes during hyperedge-to-node phase. This bidirectional flow enables nodes to receive and integrate relevant global information, facilitating timely updates to their states.

\begin{figure}
    \centering
    \includegraphics[width=1\linewidth]{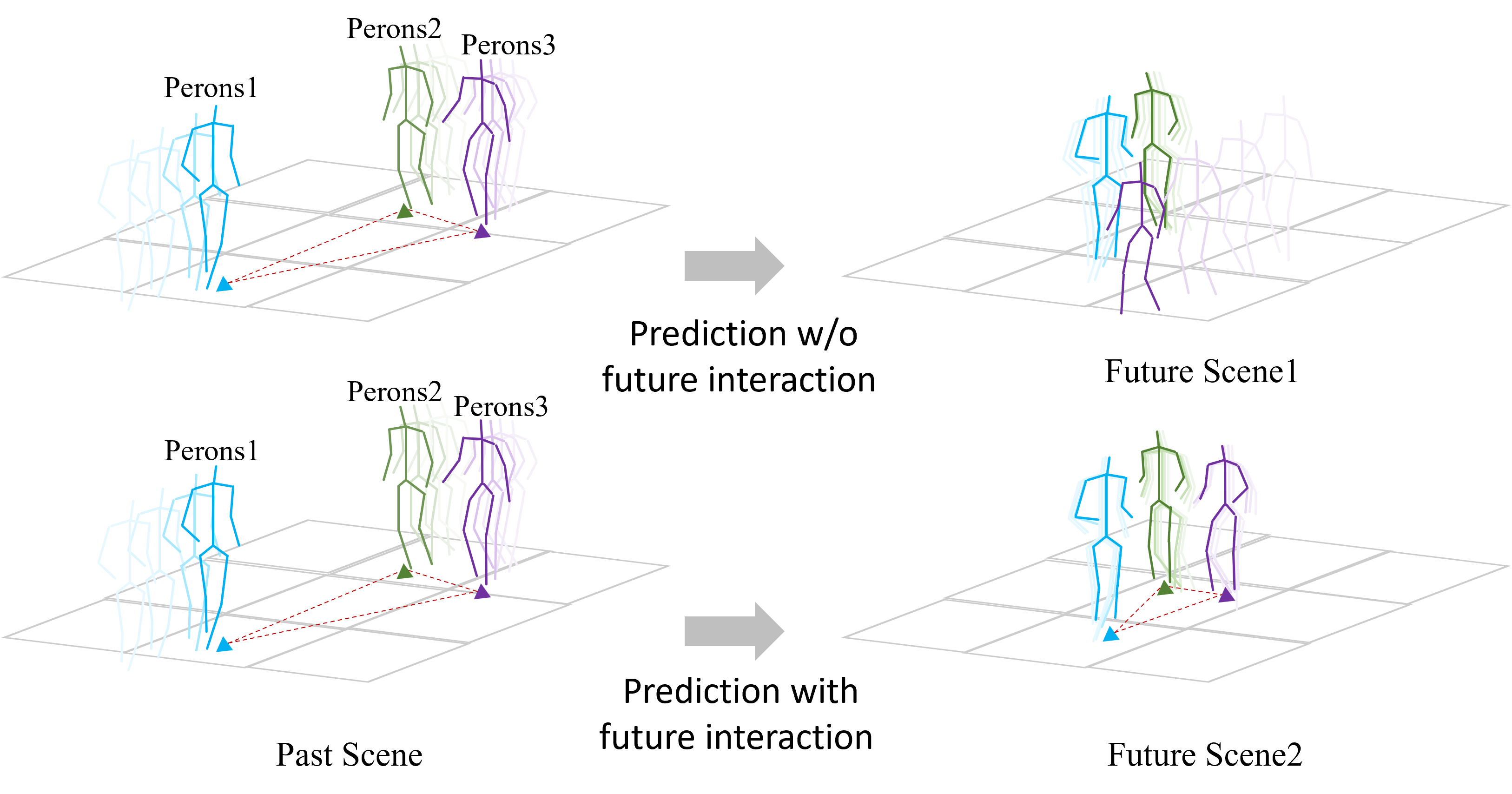}
    \caption{Illustration of our motivation. In the past scene, person 2 is walking together with person 3, while person1 is walking towards person 2. Most current methods consider the interaction during this phase. (The red dash lines denote the interaction between different individuals.) In future scene 1 and 2, a sudden situation occurs-person1 meet person 2 and  stops to talk. If we do not continually consider the existing interaction in the future, person 3 keeps walking and forgets about the person he was walking with, as shown in future scene 1. In contrast, if we think about interaction in the future, there would be a different result: person 3 should also stop and wait his partner, person 2, as shown in future scene 2. Our method is dedicated to making prediction that comply with scene 2.}
    \label{intro2}
\end{figure}

After message passing, we decode the learned features to predict future motions.
 Current methods \cite{xu2023joint,peng2023trajectory,guo2023back} often adopt non-autoregressive decoding and make prediction sequence in one time. The single-person methods benefit from this decoding strategy because each predicted frame is independent and avoid error accumulation. However, the multi-person scenes force us to reconsider this “independent strategy” because of the evolve environment and others' future motions. Let's discuss a special situation, as shown in Fig. \ref{intro2}. The past scene shows person 2 is walking together with person 3, while person 1 is walking towards person2. Future scene 1 shows the results when we don't consider the motions and intentions of others during prediction. We can see person1 and person 2 stop to talk, and person 3 ignore the sudden situation and keep walking. There would be a different result if we still consider the others' motions during the prediction, as shown in future scene 2. Person 2 also stops and waits for his partner, person 2, which is comply with real-world behavior. Relying only on historical information is insufficient for accurate predictions in an evolving environment, especially since human motion is variable. To improve the capability of handling sudden situation in the future, we include future interactions to our perspective and take account other people’s actions during the same window of time. Specifically, we introduce a predictor to continuously update interaction information and make future motion step by step, as shown in Fig. \ref{arc} (b). To generate accurate interaction, we design an inference loss function combine with position loss to supervise our train.

 Finally, we perform our experiments on multiple datasets, including 3DPW \cite{Marcard_2018_ECCV}, CMU-Mocap \cite{cmumocap2003}, MuPoTS-3D \cite{mehta2018single}, and Mix1\&Mix2 \cite{9709907}. The quantitative results demonstrate that our method achieves state-of-the-art performance across most datasets.

In summary, our contributions are as follows:
\begin{itemize}
\item We introduce UnityGraph, a novel graph structure for human motion prediction. UnityGraph formulates human motions as graph nodes, enabling the unified extraction of spatio-temporal information from nodes within a single graph. This perspective aligns with the inherent consistency and coupling of spatio-temporal features observed in the real world, thereby facilitating more accurate and natural predictions.
\item We develop a interactive decoding that considers both past and future information. Specifically, our method focuses not only on others' historical motions but also on the future. We propose a predictor to update relations and make future motion step by step during decoding. Meanwhile, we design a novel inference loss function combined with position loss to supervise the model effectively.
\end{itemize}

\section{Related work}

\subsection{Single-Person Motion Prediction.}
For single-person motion prediction, early methods are developed based on time series models \cite{ghosh2017learning,gopalakrishnan2019neural,wang2021pvred}. Wang et al. \cite{wang2021pvred} utilize pose velocities and temporal positional information to make prediction. However, it is inappropriate to consider consider human motion sequence as the time series task. Human motion involves multiple skeleton joints with spatial connection. For example, the connection between the knee and the ankle differs from that between the elbow and the shoulder. To capture spatial features, GCN-based models are widely used \cite{zhong2022spatio,dang2021msr,mao2019learning,sofianos2021space}. Mao et al. \cite{mao2019learning} propose a classic model based on GCN which takes into account both temporal smoothness and spatial dependencies among human body joints. Recently, Transformer-based methods have gained widespread attention in the field of motion prediction \cite{nie2023triplet,10064318,mao2020history,aksan2021spatio}. Yu et al. \cite{10064318} propose a cross-transformer network to register dynamic spatio-temporal information and seize its inherent coherence, which existing decoupled methods generally seldom consider. However, these methods primarily address single-person scenarios and overlook typical real-world interactions.

\subsection{Multi-Person Motion Prediction.} 
Multi-person scenes are practical and common. It also takes a new challenge for us to consider the sophisticated connection between different persons. Adeli et al. \cite{9709907} is the first group to leverage human-to-human interaction in prediction. Recently, more methods \cite{wang2021multi, vendrow2022somoformer, guo2022multi} demonstrate the effectiveness of Transformer in this task because of its ability to capture the global interactions among entire crowds. Wang et al. \cite{wang2021multi} employ a local-range Transformer to encode the motion of an individual in the sequence and a global-range Transformer to encode the motion of multiple individuals. Both encoded motion is then sent to a Transformer-based decoder to predict future motion. However, the design of the two-stream network diminishes the coupling of spatial-temporal features. Guo et al. \cite{guo2022multi} proposes a novel cross interaction
attention mechanism that exploits historical information of
both persons, and learns to predict cross dependencies between
the two pose sequences. This method primarily focuses on capturing human-to-human interactions and ignore the individual's joint relations, which are equally crucial for accurate prediction. Despite Transformer has strong ability, there is potential for further improvement. Liu et al. \cite{10194334} propose a multi-granularity learning module to capture different level interaction between individuals. Xu et al. \cite{xu2023joint} introduce physical constraints into relation learning, e.g., the different joints connected by bone show stronger associations and the joints belonging to the same individual are related. 

\subsection{Hypervariate Graph Network.} 
Hypervariate graph networks, which allow edges to connect more than two nodes, have gained prominence for modeling complex, higher-order relationships among nodes. This capability is less feasible with traditional graph networks \cite{bai2020adaptive,yu2017spatio,mateos2019connecting}. These networks have been successfully applied in diverse areas including trajectory prediction \cite{xu2022groupnet}, multivariate time series forecasting \cite{yi2024fouriergnn}, multi-modal tasks \cite{zeng2023multi} and human motion prediction \cite{li2024amhgcn}. Groupnet \cite{xu2022groupnet} utilizes a multiscale hypergraph neural network to capture high-order interactions at various scales from a group-wise perspective. FourierGNN \cite{yi2024fouriergnn} treats each series value as a graph node and uses sliding windows to create space-time fully-connected graphs. AMHGCN \cite{li2024amhgcn} utilizes the adaptive multi-level hypergraph representation to capture various dependencies among the human body. However, the AMHGCN framework, which is spatially designed with different levels of hyperedges, does not account for temporal dimensions and only addresses single-person scenarios. The hypergraphs in the aforementioned methods treat any entity (regardless of varieties and timestamps) as nodes, with hyperedges representing relationships beyond simple pairwise interactions. This allows the hypergraph to design different hyperedges to capture the spatio-temporal characteristics of the object within a single graph, ensuring the spatio-temporal coupling of the object and addressing gaps left by previous methods \cite{sofianos2021space,wu2019graph,cao2020spectral}.

\section{Methodology}

\subsection{Problem Formulation}
 Given a scene with $N$ persons, each person has $J$ skeleton joints, we define the observed sequence of the $n$-th person as ${X^{n}_{1:T} }=\left [x_{1}^{n} , x_{2}^{n},...,x_{T}^{n} \right ]  $, where $n\in  \left \{ 1,2,...,N \right \}$, $N$ denotes the number of observed people, $T$ represents the observed motion sequence length, and each $x_{t}^{n}\in \mathbb{R}^{J\times 3}$ denotes the
joints' 3-D coordinates of the $n$-th person
at the $t$-th motion sequence. Our objective is to predict the future motion sequence of the $n$-th person, denoted as ${\widehat{Y}^{n}_{T+1:T+P} }=\left[\hat{y}_{T+1}^{n} , \hat{y}_{T+2}^{n},...,\hat{y}_{T+P}^{n} \right ]  $, where $P$ denotes the predicted motion sequence length. The ground-truth of the $n$-th person can be defined as ${Y^{n}_{T+1:T+P} }=\left [ y_{T+1}^{n} , y_{T+2}^{n},...,y_{T+P}^{n} \right ]$.

\subsection{Hypergraph Construction}
\begin{figure} 
	
	\centering
	
	\includegraphics[width=0.45\textwidth]{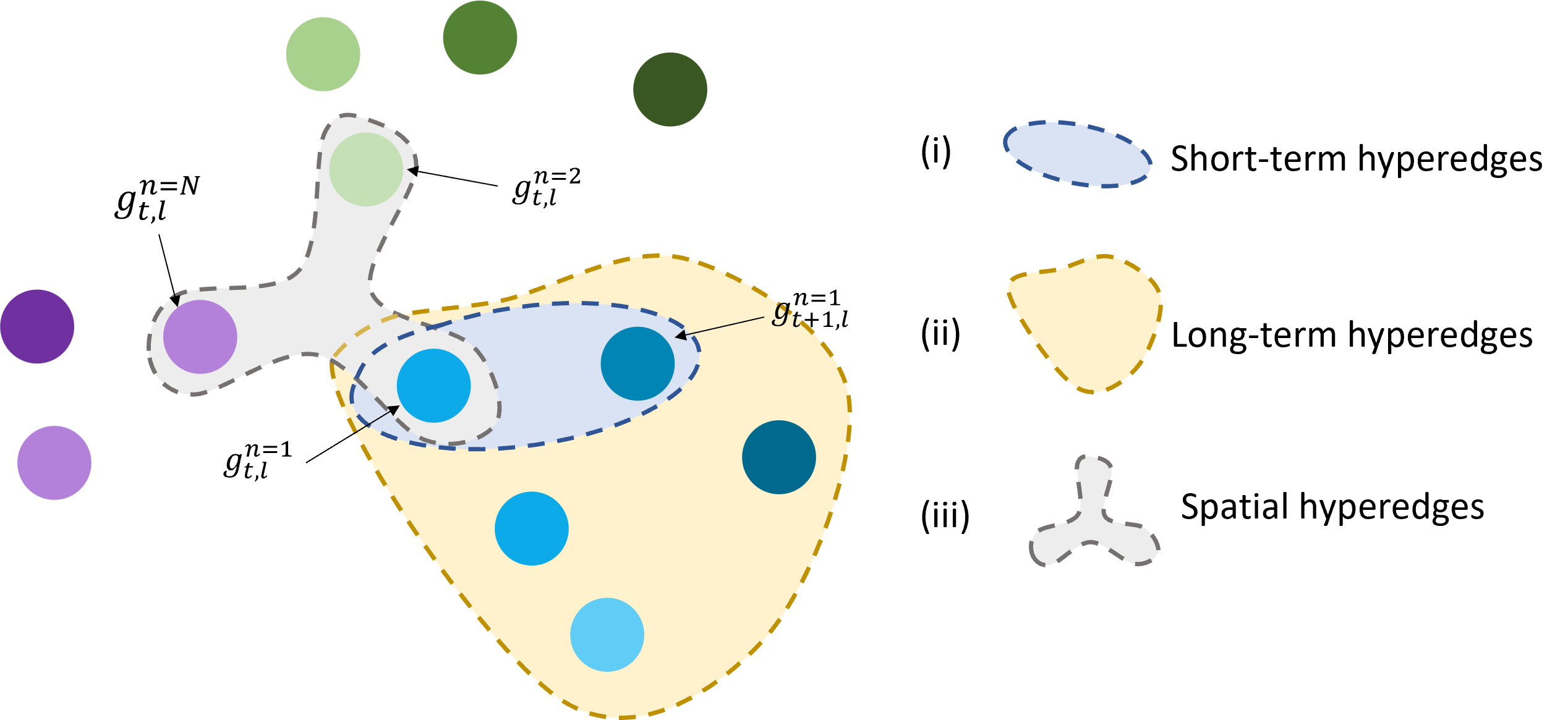}

	\caption{The illustration of nodes and hyperedges initialization. We regard the observed poses of $N$ persons as nodes of the graph and set different hyperedges to explore the relations between nodes for capture spatio-temporal features. (i) We associate the nodes of two adjacent frames with the short-term hyperedges for each individual. (ii) The long-term hyperedges consist of all nodes of time length $T$. (iii) The spatial hyperedges connect all nodes in the same frame. For clarity, some nodes and hyperedges are omitted in this figure.}
	\label{arc1}

\end{figure}

Many existing methods \cite{ 9709907,wang2021multi} adopt an intuitive approach to model spatial and temporal features independently. We argue that such separate feature learning is not sufficient to capture the complex spatial-temporal dependencies underlying the motion sequence. To comprehensively model these intrinsic composite features, we consider using a  specially designed hypergraph for its object-oriented character, see Figure \ref{arc1}. Mathematically, this hypergraph can be defined as a quaternion $\mathcal{G}=\left \{  \mathcal{V},E_{1},E_{2},E_{3}\right \}$, where $\mathcal{V}$ is a set of $N \times T$ nodes, $E_{1},E_{2}$ and $E_{3}$ are the sets of three different hyperedges.

\subsubsection{Nodes Initialization}

The nodes are composed of  features that derive from $N \times T$ poses. We adopt a graph attention network (GAT) as the encoder: 
\begin{equation}
    g _{t,l}^{n} =\textbf{GAT}\left ( x_{t}^{n},W\right ) \\\ \left (l=0\right )
\end{equation}
where $W$ is the set of parameters of the input pose attention graph and $l$ is the number of update layer and $l=0$ meant the initial stage. Accordingly, the output is $ g_{t,l}^{n}$, which
is a motion representation attending to different joint connectivity. The node of the $n$-th person at the $t$-th frame can be denotes as
$g _{t,l}^{n}\in \mathbb{R}^{J\times 3} $. Thus, the node set can be express as $\mathcal{V}=\left \{g _{t,l}^{n}|t=1,2,\cdots,T; n=1,2,\cdots,N \right \}$.

\subsubsection{Hyperedges Initialization}
To capture high-order spatio-temporal correlations among nodes, we design three types of hyperedges: short-term hyperedges, long-term hyperedges, and spatial hyperedges.

\textbf{Short-term hyperedges} 
We design $N\times\left(T-1\right)$ short-term hyperedges, connecting each person's $t$-th node to the $(t+1)$-th node. Formally, we define a short-term hyperedge as follows:
\begin{equation}
    (e_{1})_{t,l}^{n} =\textbf{Aggregate}\left ( g_{t,l}^{n}, g_{t+1,l}^{n}\right ) \\\ \left (l=0\right )
\end{equation}
Where $\textbf{Aggregate}(\cdot)$ is the mean operation. Short-term hyperedges allows the propagation of information between two neighboring nodes $g_{t,l}^{n}$ and $g_{t+1,l}^{n}$ in the temporal domain. The set of
short-term hyperedges $E_{1} $ can
be expressed as the following two steps:
\begin{equation}
     E_{1} =\bigcup_{n=1}^{N}\left \{ (e_{1})_{t,l}^{n}| t\in\left \{ 1,2,\cdots ,T-1 \right \}  \right \} \\\ \left (l=0\right )
\end{equation}

\textbf{Long-term hyperedges} 
To capture temporal features on long-term, we design $N$ long-term hyperedges. Each hyperedge connects
$T$ nodes of single person:
\begin{equation}
     (e_{2})_{l}^{n} =\textbf{Aggregate}\left ( g_{1,l}^{n}, g_{2,l}^{n},\cdots, g_{T,l}^{n}\right ) \\\ \left (l=0\right )
\end{equation}
The set of long-term hyperedges $E_{2}$ can be expressed as: 
\begin{equation}
    E_{2}=\left \{    (e_{2})_{l}^{1} ,  (e_{2})_{l}^{2} ,\cdots ,  (e_{2})_{l}^{N}   \right \}  \\\ \left (l=0\right )
\end{equation}

\textbf{Spatial hyperedges}
In addition, we define $T$ spatial hyperedges to capture the interactions across different persons. The
spatial hyperedges are defined as follows:

\begin{equation}
     (e_{3})_{t,l}  =\textbf{Aggregate}\left ( g_{t,l}^{1}, g_{t,l}^{2},\cdots ,g_{t,l}^{N}\right ) \\\ \left (l=0\right ) 
\end{equation}
Where spatial hyperedges focus on aggregating information across $N$ individuals within
$t$-th frame to obtain interaction features. The set of
temporal hyperedges can be defined as:
\begin{equation}
    E_{3}=\left \{(e_{3})_{1,l},(e_{3})_{2,l},\cdots,(e_{3})_{T,l}    \right \}  \\\ \left (l=0\right )
\end{equation}

\begin{figure} 
	
	\centering
	
	\includegraphics[width=0.5\textwidth]{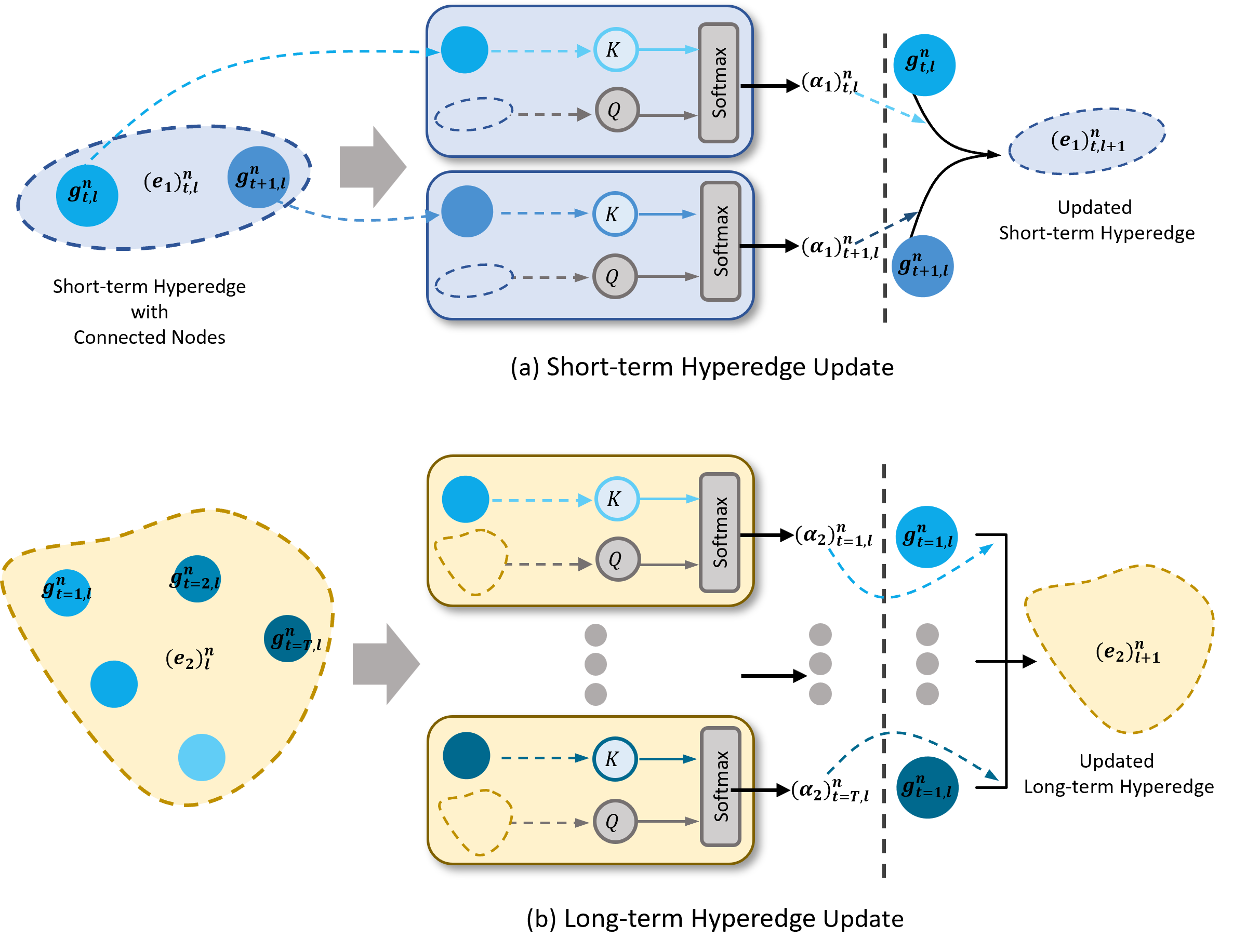}

	\caption{The illustration of short-term and long-term hyperedges update. (a) On short-term, we select node $g_{t,l}^{n}$ and its neighbor $g_{t+1,l}^{n}$ to update the edge $(e_{1})_{t,l}^{n}$ which connects them. (b) We update long-term hyperedges by aggregating all nodes from $g_{t=1,l}^{n}$ to $g_{t=T,l}^{n}$.}  
	\label{update1}

\end{figure}
\subsection{Dynamic Message Passing}
Spatial and temporal features are different in nature and have different impacts. In order to capture their own features without deconstructing the inherent correlation and coupling between them. We customize a neural message passing method to obtain the spatio-temporal features iteratively through node-to-hyperedge and hyperedge-to-node update.

\subsubsection{Node-to-hyperedge update}
To propagate short-term information through hyperedges among nodes, we deploy a multi-head attention mechanism to
compute the weight between hyperedge $(e_{1})_{t,l}^{n}$ and nodes $g_{t,l}^{n}$, $g_{t+1,l}^{n}$, as shown in Fig. \ref{update1} (a). The attention weight is computed as follows: 

\begin{equation}
    (\alpha_{1}) _{t,l}^{n}=\textbf{Softmax}\left ( \frac{\left (  (e_{1})_{t,l}^{n}\cdot W_{e,l}\right )\cdot \left (  g_{t,l}^{n}\cdot W_{g,l}\right )^{T}}{\sqrt{d_{e_{1}} }} \right )
    \label{weight}
\end{equation}
Where $(\alpha_{1}) _{t,l}^{n}$ denotes the attention weight for node 
$g_{t,l}^{n}$. $W_{e}^{l}$ and $W_{g}^{l}$ are learnable transformation
matrices and $d_{e_{1}}$ is the dimension of the $(e_{1})_{t,l}^{n}\cdot W_{e }^{l}$.

With the hypergraph attention mechanism, the updated $t$-th spatial hyperedges $(e_{1})_{t,l}^{n}$
is computed by aggregating information from the $t$-th node $g_{t,l}^{n}$ and
the $\left(t+1\right)$-th node $g_{t+1,l}^{n}$ . Specifically, the hyperedge representation is computed
by the following equation:
\begin{equation}
    (e_{1})_{t,l+1}^{n} =  \sigma \left[(\alpha_{1}) _{t,l}^{n}\cdot g_{t,l}^{n} + (\alpha_{1}) _{t+1,l}^{n}\cdot g_{t+1,l}^{n} \right]+ (e_{1})_{t,l}^{n} 
\end{equation}
Where $(e_{1})_{t,l+1}^{n}$ denotes the updated short-term hypredge of the $(l+1)$-th layer. $\sigma\left(\cdot\right)$ is an activation function.

Contrast to the short-term, the long-term hypredge update need to use all $T$ nodes of the $n$-th individual. The process can be expressed as:
\begin{equation}
        (\alpha_{2}) _{t,l}^{n}=\textbf{Softmax}\left ( \frac{\left (   (e_{2})_{l}^{n}\cdot W_{e,l}\right )\cdot \left (  g_{t,l}^{n}\cdot W_{g,l}\right )^{T}}{\sqrt{d_{e_{2}}  }} \right )
\end{equation}

\begin{equation}
    (e_{2})_{l+1}^{n} =  \sigma \left[\sum_{t=1}^{T}(\alpha_{2}) _{t,l}^{n}\cdot g_{t,l}^{n} \right]+ (e_{2})_{l}^{n} 
\end{equation}
Where $(\alpha_{2}) _{t,l}^{n}$ denotes the attention weight for node  $g_{t,l}^{n}$. $(e_{2})_{l+1}^{n}$ denotes the updated long-term hypredge of the $(l+1)$-th layer.

\begin{figure} 
	
	\centering
	
	\includegraphics[width=0.5\textwidth]{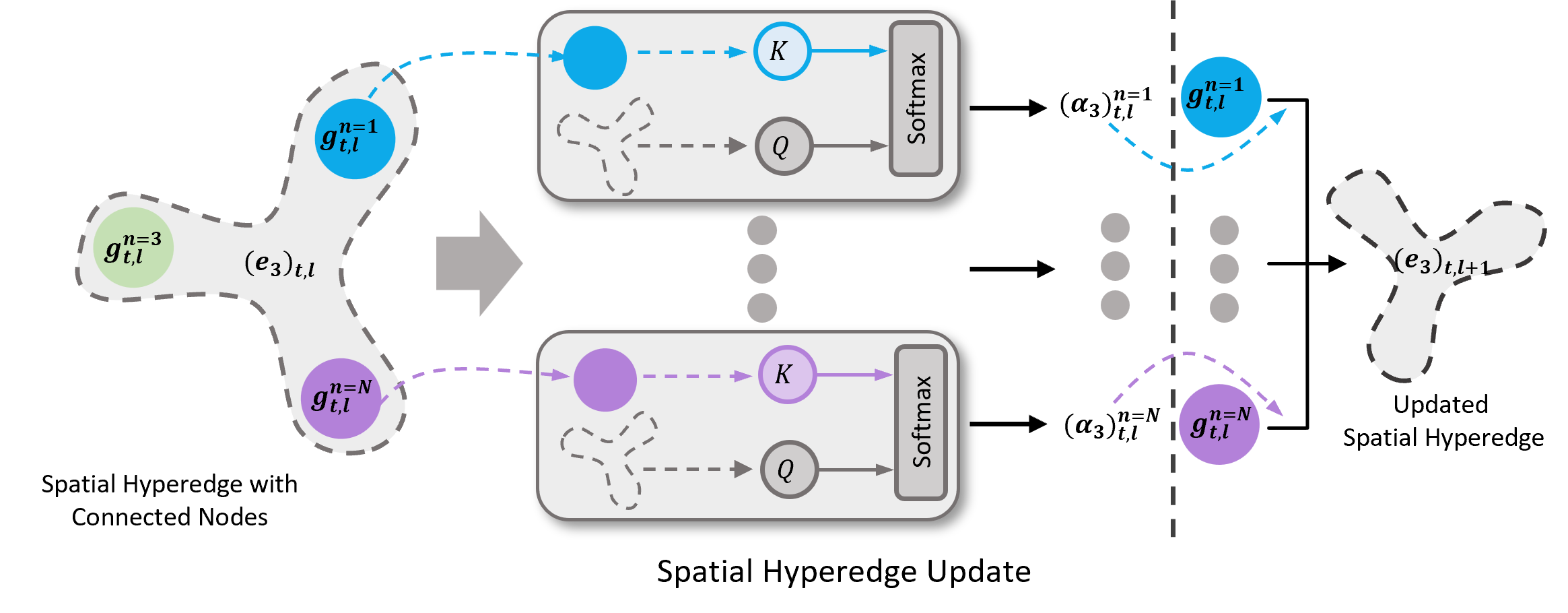}

	\caption{The process of updating a spatial hyperedge involves influence from all nodes in the same frame.}
	\label{update2}

\end{figure}

Similarly, we update spatial hyperedges by incorporating
information from connected nodes, as shown in Fig. \ref{update2}. Specifically, the updated spatial hyperedges $(e_{3})_{t,l+1} $ is computed as follows:
\begin{equation}
        (\alpha_{3}) _{t,l}^{n}=\textbf{Softmax}\left ( \frac{\left (  (e_{3})_{t,l}\cdot W_{e}^{l}\right )\cdot \left (  g_{t,l}^{n}\cdot W_{g}^{l}\right )^{T}}{\sqrt{d_{e}  }} \right )
\end{equation}

\begin{equation}
    (e_{3})_{t,l+1}  =  \sigma \left[\sum_{n=1}^{N}(\alpha_{3}) _{t,l}^{n}\cdot g_{t,l}^{n} \right]+ (e_{3})_{t,l}
\end{equation}
Where $(\alpha_{3}) _{t,l}^{n}$ denotes the attention weight of different nodes. The weight $(\alpha_{3}) _{t,l}^{n}$
allows the model to focus on the most
relevant nodes when updating the hyperedge representations. The updated spatial hyperedges $(e_{3})_{t,l+1} $ obtains the features from all individuals of $t$-th frame and captures the interactions across different individual.

\subsubsection{Hyperedge-to-node update}
\begin{figure} 
	
	\centering
	
	\includegraphics[width=0.5\textwidth]{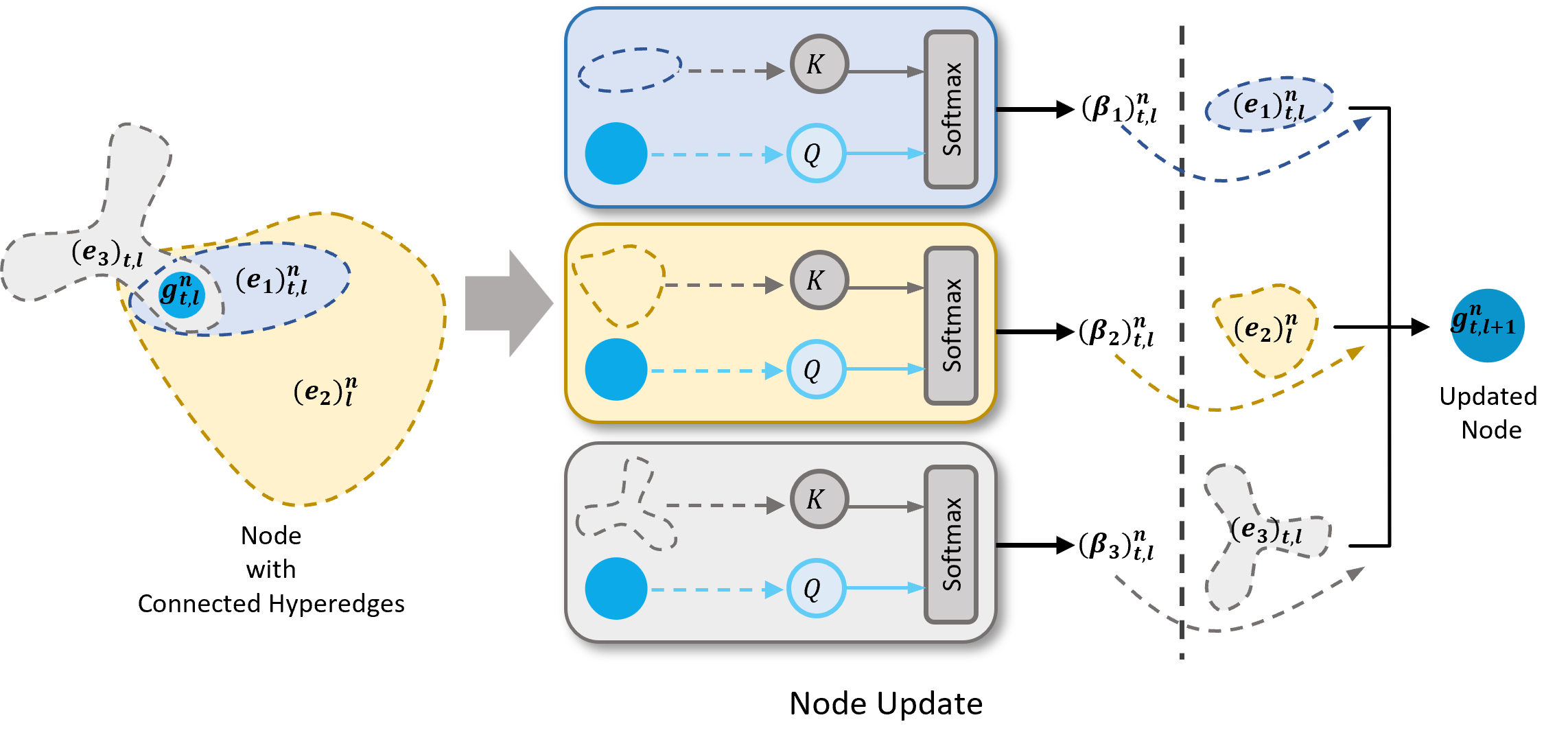}

	\caption{The illustration of a node update. The process of updating nodes involves all hyperedges associated with the node $g_{t,l}^{n}$.}
	\label{update3}

\end{figure}

In this section, we update nodes by propogating information through hyperedges, as shown in Fig. \ref{update3}. Similar to the hyperedges update, we first calculate the weight between a node $g_{t,l}^{n}$ and the temporal and spatial hyperedges $(e_{1})_{t,l}^{n},(e_{2})_{l}^{n}, (e_{3})_{t,l}$ connected
with it as follows:
\begin{equation}
   (\beta_{1})_{t,l}^{n}=\textbf{Softmax}\left ( \frac{\left (  (e_{1})_{t,l}^{n}\cdot W_{e  }^{l}\right )^{T}\cdot \left (  g_{t,l}^{n}\cdot W_{g}^{l}\right )}{\sqrt{d_{g}}} \right )
    \label{weight2}
\end{equation}

\begin{equation}
    (\beta_{2})_{t,l}^{n}=\textbf{Softmax}\left ( \frac{\left (  (e_{2})_{l}^{n}\cdot W_{e  }^{l}\right )^{T}\cdot \left (  g_{t,l}^{n}\cdot W_{g}^{l}\right )}{\sqrt{d_{g}}} \right )
\end{equation}

\begin{equation}
   (\beta_{3})_{t,l}^{n}=\textbf{Softmax}\left ( \frac{\left (  (e_{3})_{t,l}\cdot W_{e}^{l}\right )^{T}\cdot \left (  g_{t,l}^{n}\cdot W_{g}^{l}\right )}{\sqrt{d_{g}}} \right )
    \label{weight3}
\end{equation}
Where $ (\beta_{1})_{t,l}^{n}, (\beta_{2})_{t,l}^{n}, (\beta_{3})_{t,l}^{n} $ denotes the attention weight for three different hyperedges. $d_{g}$ is the dimension of the $g_{t,l}^{n}\cdot W_{g}^{l}$. 

The representation of the node in the $(l + 1)$-th layer
are computed using the following equations:

\begin{equation}
\begin{split}
    g_{t,l+1}^{n} &=  \textbf{MLP}\left((\beta_{1})_{t,l}^{n}\cdot (e_{1})_{t,l}^{n}\right) + \textbf{MLP} \left((\beta_{2})_{t,l}^{n}\cdot (e_{2})_{l}^{n}\right) \\
    &+\textbf{MLP}\left((\beta_{3})_{t,l}^{n}\cdot (e_{3})_{t,l}\right) + g_{t,l}^{n} 
    \end{split}
\end{equation}
Where $g_{t,l+1}^{n}$ is the updated representation of
node in $(l + 1)$-th layer.

Overall, we repeat the node-to-hyperedge and
hyperedge-to-node phases for $L$ times and obtain the
updated graph nodes. Finally, we get the sequences of $N$ individuals' motion representation by:
\begin{equation}
\begin{aligned}
Z^{1}_{1:T} = \mathbf{Concat}&\left ( g_{1,L}^{1} ,g_{2,L}^{1},\cdots,g_{T,L}^{1}\right ) \\
Z^{2}_{1:T}  = \mathbf{Concat}&\left ( g_{1,L}^{2} ,g_{2,L}^{2},\cdots,g_{T,L}^{2}\right ) \\
 \vdots& \\
Z^{N}_{1:T}  = \mathbf{Concat}&\left ( g_{1,L}^{N} ,g_{2,L}^{N},\cdots,g_{T,L}^{N}\right ) 
\end{aligned}
\end{equation}

\subsection{Interactive decoding}

\begin{figure} 
	
	\centering
	
	\includegraphics[width=0.5\textwidth]{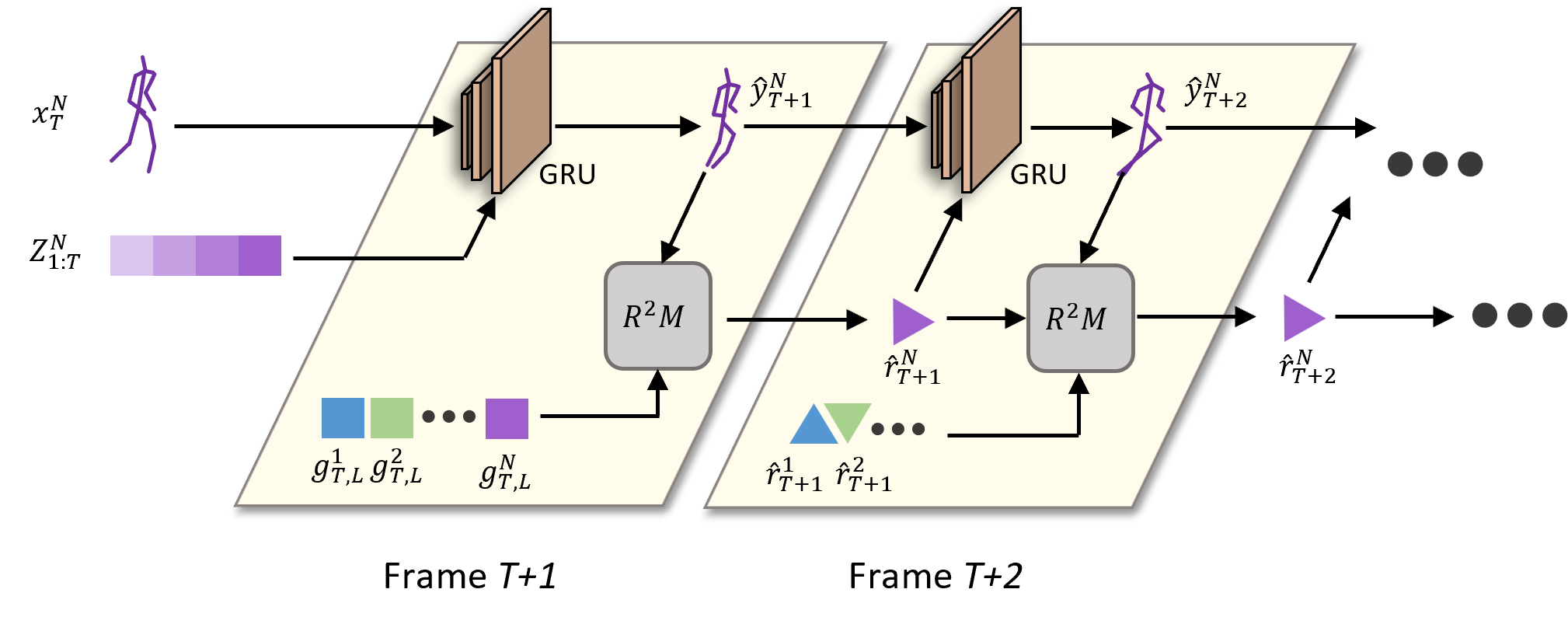}

	\caption{The illustration of the predictor in Fig. \ref{arc} (b). We take person $N$ as the example. For the first prediction frame $T+1$, the motion representation sequence of individual $N$ is used as input to the predictor. Meanwhile, to reduce the uncertainty of the predicted motion $\hat{y}_{T+1}^{N}$ and enhance accuracy, the observed motion $X_{T}^{N}$ from the last observed frame $T$ is also included as input. Then we utilize the prediction motion to update the interaction relation $\hat{r}_{T+1}^{N}$ of individual $N$ by $R^{2}M$. Subsequent, we make new prediction $\hat{y}_{T+2}^{N}$ using the previous motion $\hat{y}_{T+1}^{N}$ along with the updated relation $\hat{r}_{T+1}^{N}$. This step-by-step until frame $T+P$.}
	\label{predictor}

\end{figure}

To accurately predict future poses, relying solely on historical information is insufficient. We also consider dynamics in the future, such as others' future motions occurring during the same window of time. Formally, after updating all the historical features, we obtain motion sequences of $N$ persons. Next, we apply a predictor to generate the set of future poses recursively, as shown in Fig. \ref{predictor}. We take person $N$ as the example, mathematically the procedure at $T+1$ frame can be expressed as:
\begin{equation}
\hat{y}_{p}^{N} = \textbf{GRU} \left(Z^{N}_{1:T} , x_{T}^{N}\right)  \ \left ( p=T+1 \right )
\end{equation}
Where $p$ denotes the $p$-th predicted frame and $\hat{y}_{p}^{N}$ is the $p$-th predicted motion of $N$-th person $\left ( p\in \left ( T+1,T+P \right )  \right ) $. The first prediction motion is matter in autoregressive decoding, we take the last observed motion $x_{T}^{N}$ as input to reduce uncertainty in latent space.

The reasoning persons’ interactions $\hat{r}_{p+1}^{N}$ are obtained by the relation reasoning module $R^{2}M$:
\begin{equation}
\begin{aligned}   
\hat{r}_{p}^{N}=&\mathbf{{R^{2}M }}\left ( \bigcup_{n=1}^{N}g_{T,L}^{n},\hat{y}_{p}^{N}\right ) \ \left ( p=T+1 \right ) \\
= &\sum_{m=1}^{N-1}I_{T}^{\left ( N,m \right )}+ \hat{y}_{p}^{N}
\end{aligned}
\end{equation}
Where $m$ denotes a distinct individual from person $N$. $I_{p}^{\left ( N,m\right )}$ is the interaction correlation between the person $N$ person and the person $m$, it can be calculated as follow:
\begin{equation}
    I_{T}^{\left ( N,m \right )}= \textbf{ATT} \left(g_{T,L}^{N}, g_{T,L}^{m} \right)
\end{equation}

Where $\mathbf{ATT}\left(\cdot\right)$ is the calculation of attention score. Subsequently, when $p\ge  T+1 $, the computation can be expressed as follows :
\begin{equation}
    \hat{y}_{p}^{n} = \textbf{GRU}\left(\hat{r}_{p-1}^{N}, \hat{y}_{p-1}^{N}\right) \ \left ( p>  T+1 \right )
\end{equation}

\begin{equation}
\begin{aligned}   
\hat{r}_{p}^{N}=&\mathbf{{R^{2}M }}\left ( \bigcup_{n=1}^{N}\hat{r}_{p-1}^{n},\hat{y}_{p}^{N}\right ) \ \left (p>  T+1\right ) \\
= &\sum_{m=1}^{N-1}I_{p-1}^{\left ( N,m \right )}+ \hat{y}_{p}^{N}  
\end{aligned}
\label{eq23}
\end{equation}

\begin{equation}
    I_{p-1}^{\left ( N,m \right )}= \textbf{ATT} \left(\hat{r}_{p-1}^{N},\hat{r}_{p-1}^{m} \right)
\end{equation}

Finally, we obatain the predicted motion sequences of $N$ persons:
\begin{equation}
\begin{aligned}
\widehat{Y}^{1}_{T+1:T+P} = \mathbf{Concat}&\left ( \hat{y}_{T+1}^{1} ,\hat{y}_{T+2}^{1},\cdots,\hat{y}_{T+P}^{1}\right ) \\
\widehat{Y}^{2}_{T+1:T+P} = \mathbf{Concat}&\left ( \hat{y}_{T+1}^{2} ,\hat{y}_{T+2}^{2},\cdots,\hat{y}_{T+P}^{2}\right ) \\
 \vdots& \\
\widehat{Y}^{N}_{T+1:T+P} = \mathbf{Concat}&\left ( \hat{y}_{T+1}^{N} ,\hat{y}_{T+2}^{N},\cdots,\hat{y}_{T+P}^{N}\right ) \\
\end{aligned}
\end{equation}

\subsection{Loss Function}
We design a joint loss function to evaluate proposed network. The loss function
includes three parts, prediction loss, reconstruction loss,
and inference loss.

The prediction loss $\mathcal{L}_{pre}$ and reconstruction loss  $\mathcal{L}_{rec}$ are used to measure the
error between the predicted motion and the corresponding ground truth. The calculations are defined as follows:
\begin{equation}
    \mathcal{L}_{pre}=\left \| {\widehat{Y}^{n}_{T+1:T+P} }-Y^{n}_{T+1:T+P}  \right \|_{2}
\end{equation}

\begin{equation}
    \mathcal{L}_{rec}=\left \| {\widehat{X}^{n}_{1:T} }-X^{n}_{1:T}   \right \|_{2}
\end{equation}
Where motion sequence $\widehat{X}^{n}_{1:T}$ is the reconstructed output of $X^{n}_{1:T}$ from our network. $ \left \| \cdot  \right \|_{2}$ denotes the $\ell_{2}$ norm. 

In addition, we design the inference loss $\mathcal{L}_{inf}$ to 
supervise individuals' representation reasoning:
\begin{equation}
\mathcal{L}_{inf}=\sum_{n=1}^{N}\sum_{p=T+1}^{T+P}\left \| \hat{r}_{p}^{n}- r_{p}^{n}\right \|_{2}
\end{equation}
Where the ground truth $r_{p}^{n}$ is calculated by the equation.\ref{eq23} with the input $Y^{n}_{T+1:T+P}$. 

The overall loss function of UnityGraph is a weighted sum of the
three loss terms: 
\begin{equation}
    \mathcal{L} =  \lambda _{pre}\mathcal{L}_{pre}+\lambda_{rec}\mathcal{L}_{rec} +\lambda_{inf}\mathcal{L}_{inf}
\end{equation}
Where $\lambda _{pre}, \lambda_{rec}$, and $\lambda_{inf}$ are hyperparameters that control the relative
importance of each loss term.

\begin{table*}[t]
\centering
\caption{Experimental results in MPJPE on CMU-Mocap,
MuPoTS-3D, Mix1,and Mix2 test sets. The best results are highlighted in
bold. "*" indicates that the data of Mix1\&Mix2 are not given in the paper and we reproduced with paper's code. "-" indicates that the
data are not given in the paper and the paper does not discuss situations involving more than three individuals.}
\begin{tabular}{llccc|ccc|ccc|ccc}
\hline
\multicolumn{2}{c|}{\multirow{2}{*}{Methods}}    & \multicolumn{3}{c|}{\begin{tabular}[c]{@{}c@{}}CMU-Mocap\\ (3 persons)\end{tabular}} & \multicolumn{3}{c|}{\begin{tabular}[c]{@{}c@{}}MuPoTS-3D\\ (2$\sim$3 persons)\end{tabular}}& \multicolumn{3}{c|}{\begin{tabular}[c]{@{}c@{}}Mix1\\ (9$\sim$15 persons)\end{tabular}} & \multicolumn{3}{c}{\begin{tabular}[c]{@{}c@{}}Mix2\\ (11 persons)\end{tabular}}
\\ \cline{3-14} 
\multicolumn{2}{c|}{}                            & 1sec     & 2sec     & 3sec     & 1sec     & 2sec     & 3sec  & 1sec     & 2sec     & 3sec   & 1sec     & 2sec     & 3sec\\ \hline
\multicolumn{2}{c|}{LTD \cite{mao2019learning}}                        & 13.7     & 21.9     & 32.6     & 11.9     & 18.1     & 23.4    & 21.0     & 31.9     & 41.5 & 17.2     & 25.8     & 34.5\\
\multicolumn{2}{c|}{HRI \cite{mao2020history}}                        & 14.9     & 26.0     & 30.7     & 9.4      & 16.8     & 22.9    & 18.0      & 31.4     & 42.1  & 16.0      & 27.1     & 36.7 \\
\multicolumn{2}{c|}{MRT \cite{wang2021multi}}                        & 9.6      & 15.7     & 21.8     & 8.9     & 15.9     & 22.2    & 17.3     & 29.9     & 39.7  & 12.9     & 20.9     & 28.2 \\
\multicolumn{2}{c|}{TCD* \cite{saadatnejad2023generic}}                        & 10.2      & 16.1     & 19.5     & 9.0     & 15.8    & 21.7    & 16.8    & 28.7    & 37.7  & 12.1     & 19.2    & 26.7 \\
\multicolumn{2}{c|}{TBIFormer* \cite{peng2023trajectory}}                        & 8.0      & 13.4    & 19.0     & 8.7     & 15.1     & 20.9    & 14.0     & 24.4     & 31.2  & 12.3     & 18.2     & 26.1 \\

\multicolumn{2}{c|}{JRT \cite{xu2023joint}} & 8.3      & 13.9     & 18.5     & 8.9      & 15.5     & 21.3    
 & -      & -     & -   & -      & -     & -  \\ \hline
\multicolumn{2}{c|}{Ours}                       &    \textbf{7.8}      &  \textbf{13.0}        &  \textbf{17.3}        &   \textbf{8.4}    &  \textbf{14.7}        &  \textbf{20.1}     &   \textbf{13.6}    &  \textbf{23.6}        &  \textbf{29.4}  
&   \textbf{11.0}    &  \textbf{16.2}        &  \textbf{23.8} \\ \hline
\end{tabular}
\label{cmu/mopots}
\end{table*}

\begin{table}[t]
\centering
\caption{Experimental results in VIM on 3DPW test sets. The best results are highlighted in bold. "*" indicates that the data are not given in the paper
and we reproduced with paper’s code.}

\begin{tabular}{cl|cccccc}
\hline
\multicolumn{2}{c|}{\multirow{2}{*}{Method}}                               & \multicolumn{6}{c}{\begin{tabular}[c]{@{}c@{}}3DPW\\ (2 persons)\end{tabular}}               \\ \cline{3-8}           
\multicolumn{2}{c|}{}                        & AVG           & 100          & 240           & 500           & 640           & 900                \\ \hline
\multicolumn{2}{c|}{LTD \cite{mao2019learning}}                     & 76.7          & 22.0         & 41.1          & 81.0          & 100.2         & 139.7                 \\
\multicolumn{2}{c|}{TRiPOD \cite{9709907}}                   & 84.2          & 31.0         & 50.8          & 84.7          & 104.1          & 150.4            \\
\multicolumn{2}{c|}{DViTA \cite{parsaeifard2021learning}}                   & 65.7          & 19.5         & 36.9          & 68.3          & 85.5          & 118.2                \\

\multicolumn{2}{c|}{MRT \cite{wang2021multi}}                     & 59.2          & 21.8         & 39.1          & 65.1          & 75.9          & 94.1               \\
\multicolumn{2}{c|}{TCD* \cite{saadatnejad2023generic}}              & 51.0 &10.8 & 24.8 & 55.1          & 68.3 & 96.1        \\
\multicolumn{2}{c|}{TBIFormer*\cite{peng2023trajectory}}              & 48.4 &9.8 & 22.5 & 50.6          & 63.1 & 94.2        \\
\multicolumn{2}{c|}{JRT \cite{xu2023joint}}                     & 47.2          & 9.5          & 22.1          & 48.7          & 62.8          & 92.8              
\\
\hline

\multicolumn{2}{c|}{Ours}          & \textbf{46.7}          & \textbf{9.4}         & \textbf{21.7}          & \textbf{47.6} & \textbf{62.5}          & \textbf{92.5}         \\ \hline
\label{3dpw-vim}
\end{tabular}

\end{table}

\section{Experiment and discussions}

\subsection{Baselines}
To assess the effectiveness of our method, we select it against a variety of baselines, including classical approaches such as LTD \cite{mao2019learning}, TRiPOD \cite{9709907}, DViTA \cite{parsaeifard2021learning}, and MRT \cite{wang2021multi}, in addition to the single-person motion prediction method TCD \cite{saadatnejad2023generic}. We also incorporate state-of-the-art models, JRT \cite{xu2023joint} and TBIFormer \cite{peng2023trajectory}, as baselines. LTD \cite{mao2019learning} serves as a foundational model in human motion prediction, encoding temporal within trajectory space. TRiPOD \cite{9709907} models interactions between humans and objects. DViTA \cite{parsaeifard2021learning} addresses multi-person prediction by dividing it into several single-person prediction tasks. MRT \cite{wang2021multi} highlights interaction significance in multi-person prediction tasks via a multi-range Transformer. JRT \cite{xu2023joint} investigates physical and interaction relations to the human body, achieving notable results. TBIFormer \cite{peng2023trajectory} focuses on skeleton dynamics, converting pose into multi parts to learn dynamic body part interactions.

\begin{figure*} 
	
	\centering
	
	\includegraphics[width=0.7\textwidth]{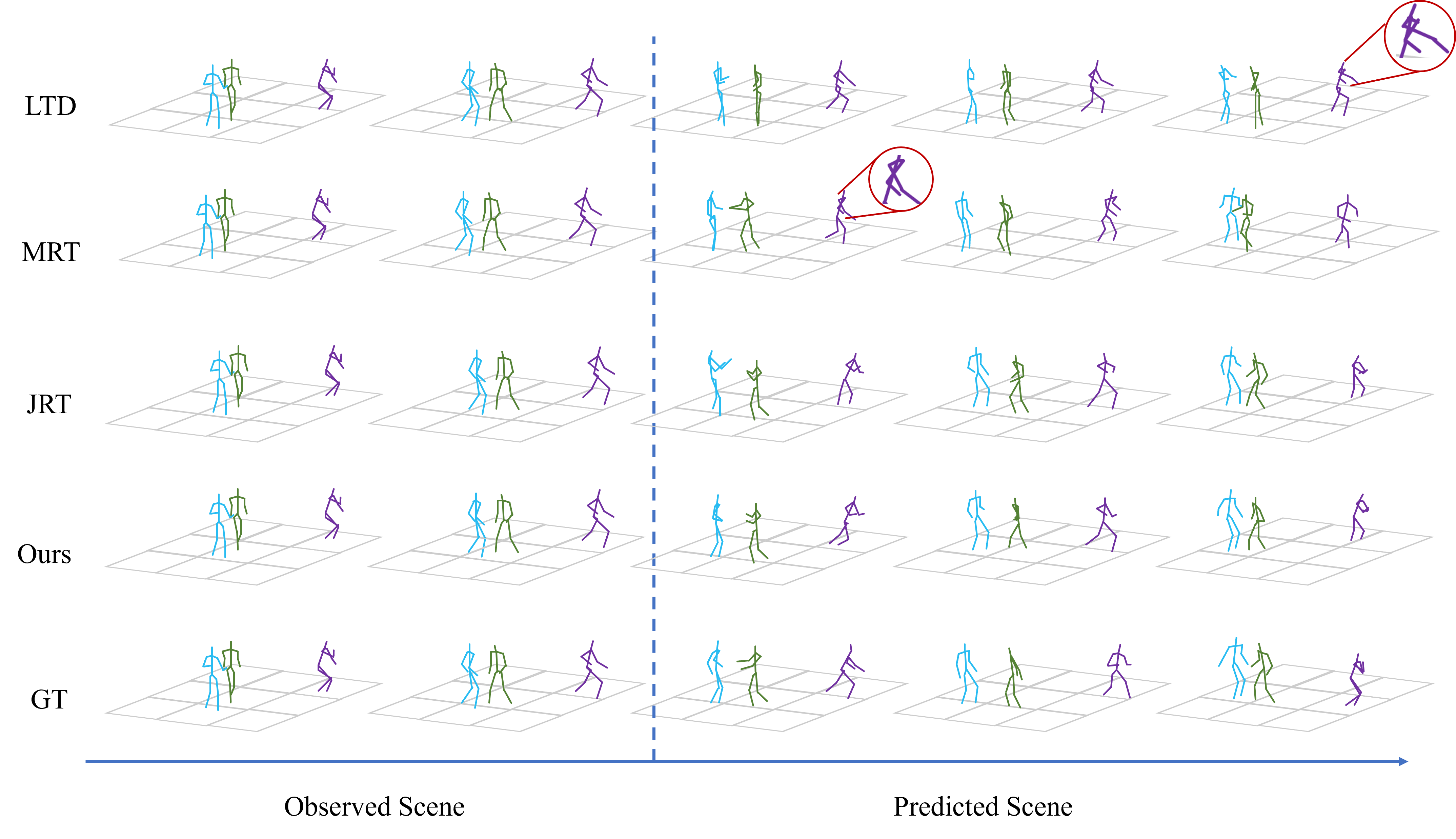}

	\caption{Visualization comparison on CMU-Mocap dataset. We compare the prediction by our method and three previous methods. Our method generates a
more natural and accurate motion prediction.}
	\label{visual1}

\end{figure*}

\begin{figure} 
	
	\centering
	
	\includegraphics[width=0.4\textwidth]{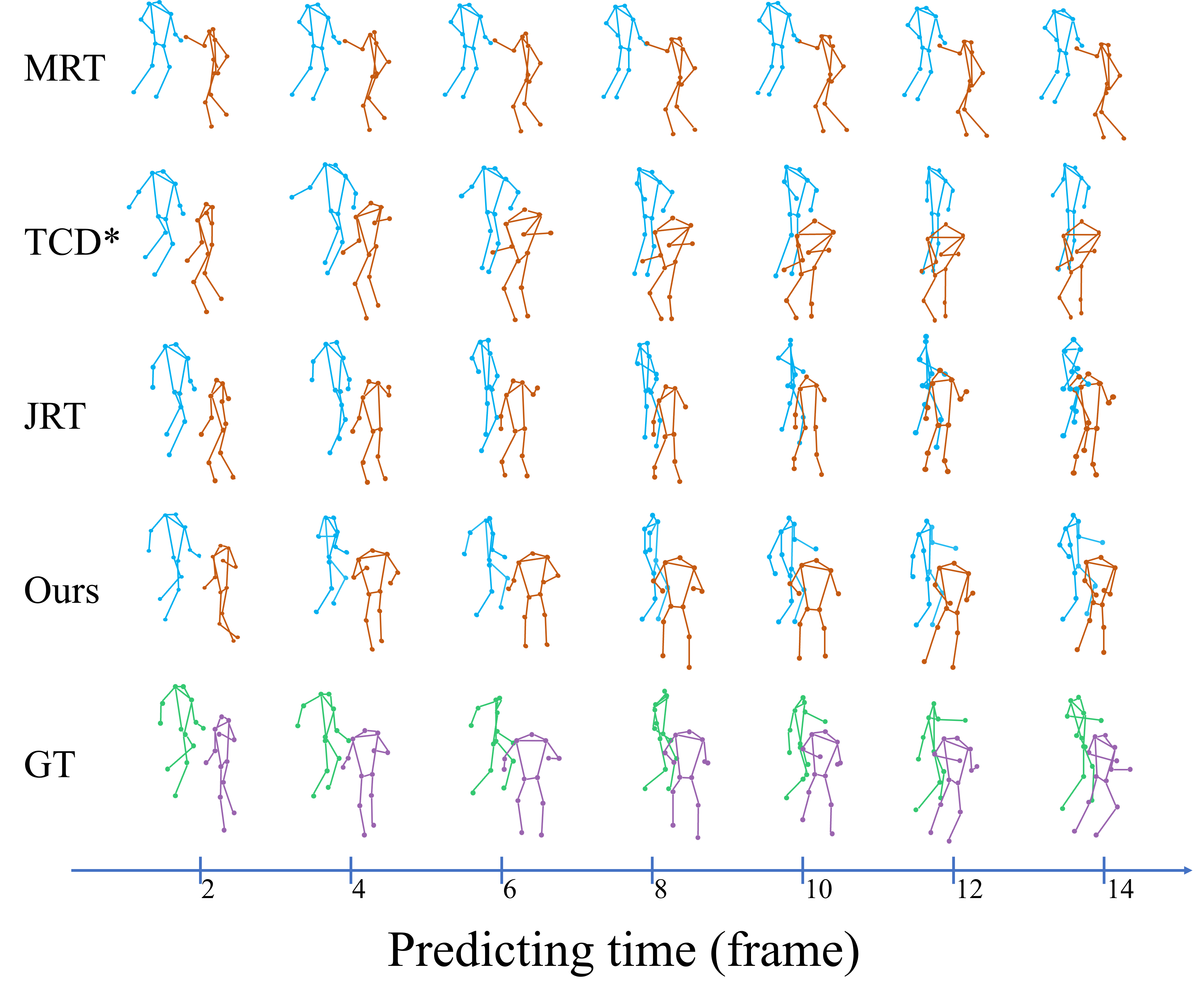}

	\caption{Visualization comparison on 3DPW dataset. We compare the prediction by our method and three previous methods. Our method generates a
more precise motion prediction.}
	\label{visual2}

\end{figure}
\subsection{Datasets}
We utilize four multi-person motion prediction benchmarks
in our experiments: CMU-Mocap \cite{cmumocap2003}, MuPoTs-3D \cite{mehta2018single}, 3DPW \cite{Marcard_2018_ECCV}, and a Mixed dataset: Mix1\&Mix2. The details of these datasets are presented below.

\noindent\subsubsection{CMU-Mocap \cite{cmumocap2003}} The Carnegie Mellon University Motion Capture Database (CMU-Mocap) collects data from 112 subjects. Most scenes capture the movements of one person, and only 9 scenes include the movements and interactions of two persons. We follow study \cite{wang2021multi} to combine samples from both one-person and two-person scenes to create new sequences including three individuals. We aim to predict future 3000ms (45 frames) motion using the historical 1000ms (15 frames) motion.

\noindent\subsubsection{MuPoTS-3D \cite{mehta2018single}} Multiperson Pose Test Set in 3D (MuPoTS-3D) consists of over 8000 frames collected from 20 sequences with 8 subjects. Following previous
works \cite{wang2021multi,xu2023joint}, we evaluate our model’s performance with
the same segment length as CMU-Mocap on the test set.

\noindent\subsubsection{3DPW \cite{Marcard_2018_ECCV}}
3D Poses in the Wild (3DPW) dataset is a comprehensive 3D motion dataset acquired using mobile phone cameras. It consists of 60 videos and approximately 68,000 frames, including a variety of scenarios and motions. In our study, we adhere to the settings used in \cite{adeli2020socially,9709907,10194334,xu2023joint}. Each scene involves two individuals, and our objective is to predict the motion for the next 900 ms (14 frames) based on the historical motion of 1030 ms (16 frames).

\subsubsection{Mix1\& Mix2} In order to evaluate our method in scenarios involving a larger number of
individuals, we adopt the methodology presented in the MRT \cite{wang2021multi}. We sample data from the CMU-Mocap and Panoptic \cite{joo2016panoptic} datasets to generate the Mix1 training set. This
training set contains approximately 3,000 samples, each featuring
9 to 15 people in the scene.
Next, we combine CMU-Mocap, MuPoTS-3D and 3DPW data, namely Mix2. There are 11 persons in each scene of its 400 samples.

\subsection{Implementation Details} 
In practice, for the 3DPW dataset, we set the input length $T$ = 16, the output
length $P$ = 14, and the number of person $N$ = 2. For the CMU-Mocap and MuPoTS-3D datasets, we set $T$ = 15, $P$ = 45 and $N$ = 3.

In our model, all the MLPs
have 2 layers with the ReLU activation function, and we conducted an ablation study for
the number of iterations $L$ for the hypergraph update ranging from 1
to 10, and determined the optimal value to be 3. In the loss function, we set $\lambda _{pre} = 0.7, \lambda_{rec} = 0.2$, and $\lambda_{inf} = 0.1$. In addition, we pre-train the model on the AMASS \cite{DBLP:journals/corr/abs-1904-03278} dataset following
previous works \cite{xu2023joint,vendrow2022somoformer}, which provides massive motion sequences.
We utilize the PyTorch deep learning framework to develop
our models and optimize the training with AdamW \cite{loshchilov2017decoupled} optimizer. The
learning rate is set to $1\times 10^{-5}$ for both pre-train and finetune and decay by 0.8 every
10 epochs. The batch size is set to 256 for pre-train, 128 for finetune. The training is performed on an
NVIDIA 3080Ti GPU for 100 epochs.
\subsection{Metrics}

\subsubsection{MPJPE \cite{6682899}} Mean Per Joint Position Error (MPJPE) is a commonly used metric in human motion prediction, which calculates the average Euclidean distance between the prediction and the ground truth
of all joints. We use this metric on CMU-Mocap, MuPoTS-3D, Mix1, and Mix2 datasets.

\begin{equation}
    \mathbf{MPJPE}=\frac{1}{P}\frac{1}{N}\frac{1}{J}\sum_{t=1}^{p}\sum_{n=1}^{N}\sum_{j=1}^{J}\left \| Y_{nj}^{t}- \widehat{Y}_{nj}^{t}\right \|_{2}
\end{equation}

\subsubsection{VIM \cite{9709907}} We adopt the Visibility-Ignored Metric (VIM) to measure the displacement in the joint vector, which has a dimension of $J\times3$, following previous works. This metric is used on the 3DPW dataset.

\begin{equation}
   \mathbf{VIM@t}=\frac{1}{N}\sum_{n=1}^{N}\sqrt{\sum_{j=1}^{J}(Y_{nj}^{t}-\widehat{Y}_{nj}^{t})}
\end{equation}

\begin{figure*} 
	
	\centering
	
	\includegraphics[width=1\textwidth]{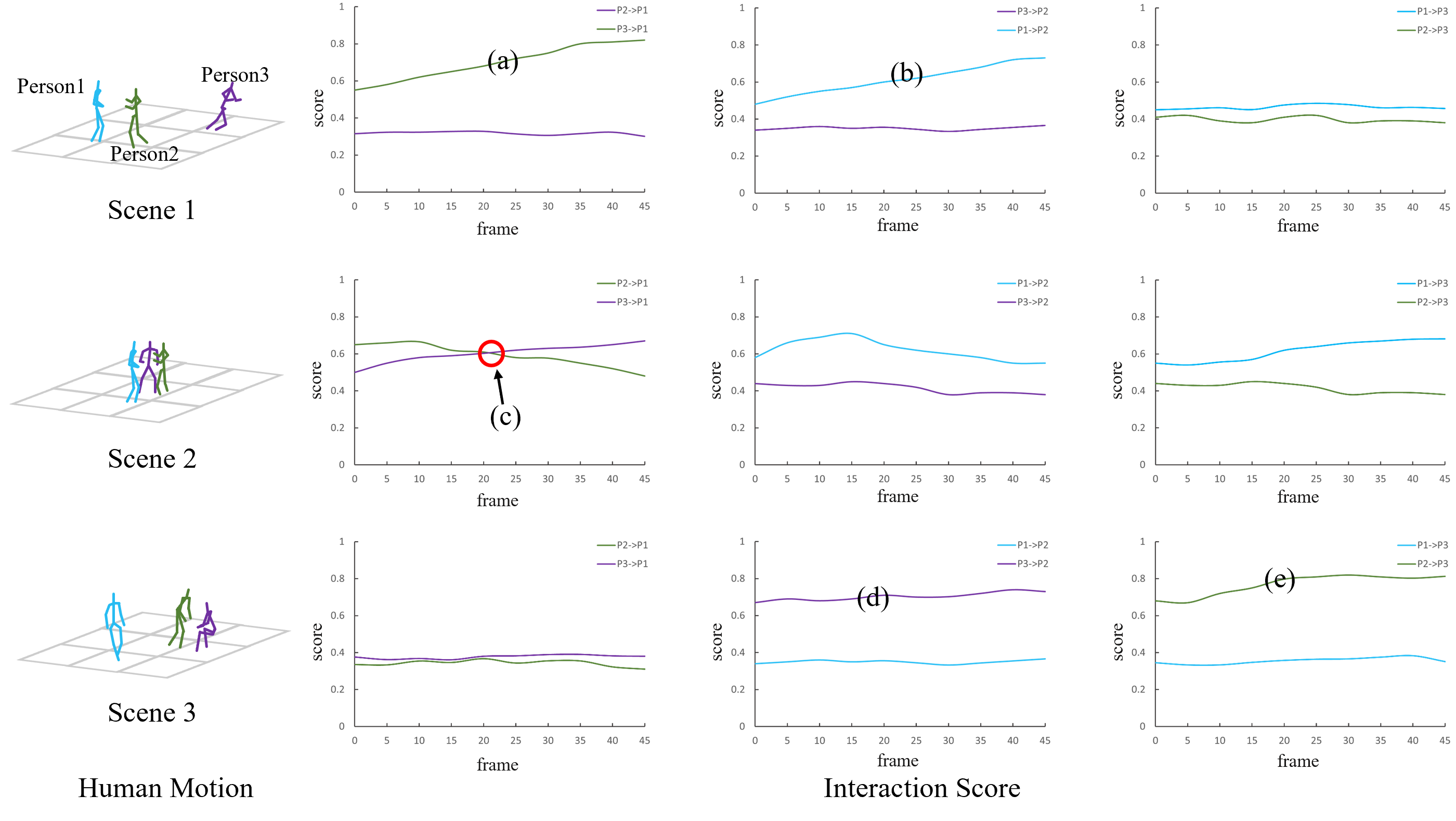}

	\caption{Visualization of interactive decoding on CMU-Mocap dataset. The figure includes three examples. For each instance, the leftmost represents the dynamics of the persons in
the sequences, and the three images on the right represent the changes in the person-level interaction scores between different persons over time, for example,
P2 → P1, P3 → P1 represent the influence of person 2 and person 3 on person 1, respectively. Better viewing in color mode.}
	\label{visual5}
\end{figure*}

\begin{figure*} 
	
	\centering
	
	\includegraphics[width=1\textwidth]{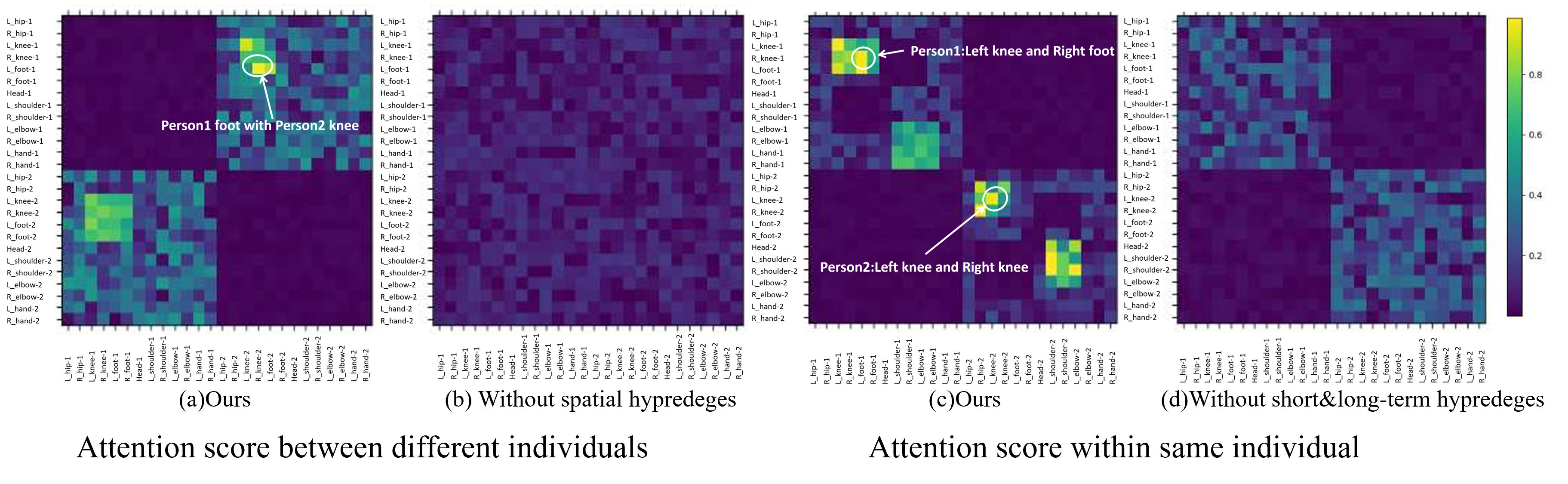}

	\caption{Attention score visualization on 3DPW dataset. We compare our method and baseline without hyperedges. Figures (a) and (b) show the attention score between different individuals, figures (c) and (d) display the attention score within same individual. (a)Our method which adopts spatial hyperedges. (b)Baseline without spatial hyperedges. (c)Our method which adopts short\&long-term hyperedges. (d)Baseline without short\&long-term hypredege. Lighter colors indicate higher attention scores and higher connections.}
	\label{visual3}
\end{figure*}

\subsection{Quantitative Results}
\subsubsection{Results on CMUMocap, MuPoTS3D, Mix1, and Mix2} For a fair comparison, we follow the same MPJPE criterion as established in previous works \cite{xu2023joint,wang2021multi,mao2019learning,peng2023trajectory,saadatnejad2023generic,mao2020history}. The experimental results in MPJPE on the several datasets are shown in Table.\ref{cmu/mopots}, where our method achieves the best performance on each dataset. Compared
to the current state-of-the-art method Joint-Relation Transformer \cite{xu2023joint}, our method reduces the MPJPE from 18.5 to 17.3 at 3s on CMU-Mocap dataset and from 21.3 to 20.1 at 3s on MuPoTS3D dataset. These improvements demonstrate the effectiveness of our method. Traditional single-person motion prediction methods HRI \cite{mao2020history} and LTD \cite{mao2019learning} focus on trajectory and ignores human interaction in multi-person motion prediction, so their experimental results are uncompetitive  compared to subsequent methods. MRT \cite{wang2021multi} utilizes a dual-path model to capture spatio-temporal features separately, it employs a simple Transformer decoder rather than designing a fusion module to aggregate different types of features. TCD
\cite{saadatnejad2023generic} is designed for single-person motion prediction and lacks the capability to seize human-to-human interactions. Consequently, its performance in multi-person scene is not as impressive as in single-person. TBIFormer \cite{peng2023trajectory} excels in short-term (1 second) forecasting but is less effective for long-term predictions. However, existing methods are limited in their ability to learn spatio-temporal features jointly and require additional steps to align these distinct features. UnityGraph achieves state-of-the-art performance since it can comprehensively learn both spatio-temporal features simultaneously and attend to the interaction-varying dependencies within the group.

\subsubsection{Results on 3DPW} We also compare the results in VIM on the 3DPW dataset between our method and several other approaches. Compared to the current state-of-the-art method, JRT\cite{xu2023joint}, our method reduces the VIM on AVG from 39.5 to 39.1. We provide 16 observed frames as input to predict the subsequent 14 frames and report the VIM in the future from 100ms to 900ms, as shown in Table. \ref{3dpw-vim}. Our method achieves state-of-the-art results, demonstrating its accuracy and strong generalization capabilities.


\subsection{Qualitative Results}
\subsubsection{Visualization of prediction result}
We provide a qualitative comparison on 3DPW test set between our method and other recent
methods, including MRT \cite{wang2021multi}, TCD \cite{saadatnejad2023generic} and JRT \cite{xu2023joint}, as shown in Fig. \ref{visual1}. Compared with the previous methods, our results are more natural and closer to the ground truth, particularly in the movement of the lower limbs. We also provide the visualization results on CMU-Mocap dataset to verify our method's effectiveness on scenes of 3 persons as shown in Fig. \ref{visual2}. We can notice that both LTD \cite{mao2019learning} and MRT \cite{wang2021multi} generate unnatural arm distortions that do not appear in our approach, as shown in the red circles. The visualization results on different datasets demonstrate the generalization and accuracy of our method.


\subsubsection{Visualization of interactive decoding}
To verify the effectiveness of reasoning relations during decoding. We visualize the interaction score from three scenes, as shown in Fig. \ref{visual5}. For each scene, we visualize each person’s interaction scores with others over time. The interaction scores in calculated by:
\begin{equation}
\begin{aligned}
\mathbf{PPC\left(P_{1},P_{2}\right)} = & \frac{\sum_{t=1}^{T} \left( \mathbf{P_{1}}\left(t,j\right) - \overline{P}_{1}\left(j\right) \right)} {\sqrt{\sum_{t=1}^{T} \left( \mathbf{P_{1}}\left(t,j\right) - \overline{P}_{1}\left(j\right) \right)^{2}}} \\
   & \times \frac{\sum_{t=1}^{T} \left( \mathbf{P_{2}}\left(t,j\right) - \overline{P}_{2}\left(j\right) \right)} {\sqrt{\sum_{t=1}^{T} \left( \mathbf{P_{2}}\left(t,j\right) - \overline{P}_{2}\left(j\right) \right)^{2}}}
\end{aligned}
\end{equation}
Where $\mathbf{P_{n}}$ denotes the motion representation of $n$-th person. $t$ denotes the $t$-th motion sequence and $j$ denotes the $j$-th skeleton joint. $\overline{P}_{n}$ denotes the mean value of person n. In scene 1, person 2 is gradually approaching person 1. It can be seen
that the interaction scores “P2→P1” and “P1 → P2” tend to
increase significantly, as shown in lines (a) and (b). In scene 2, person 1 first approaches person 2, then moves towards person 3. So the interaction scores “P3 → P1” gradually increase and “P3 → P1” are fade, as shown in the intersection (c). In scene 3, person 2 is talking to seated person 3, with no noticeable change in position. Thus the interaction scores “P3 → P2” and “P2 → P3” are both around 0.8, as shown in lines (d) and (e). The results of the three scenes above collectively show that the reasoning interactions are consistent with our intuition and real world.

\subsubsection{Visualization of attention scores}
To vividly demonstrate the effectiveness of our hyperedges, we visualize the learned attention matrices in the first layer, as shown in Fig. \ref{visual3}. We observe that in Fig. \ref{visual3} (a) the values between" left foot and right foot" for person 1 and "right foot and knee" for person 2 are significantly higher compared to those in Fig. \ref{visual3} (b), which lacks joint relations learning (The lighter the color, the higher the value). The phenomenon convinces spatial hyperedges are effective in capturing interactions between different individuals. In Fig. \ref{visual3} (c) and (d), we find that our model with short\&long-term hyperedges can effectively distinguish the highly connected joints within an individual. This is particularly noticeable in the joints that exhibit large movements when a person performs an action, such as boxing and playing basketball. In summary, it’s evident that our method can more explicitly represent different relations.

\begin{table}[]
\centering
\caption{Computational complexity analysis on CMU-Mocap dataset based on MPJPE. $L$ denotes the number of iteration.}
\begin{tabular}{c|c|c|ccc}
\hline
\multirow{2}{*}{\begin{tabular}[c]{@{}c@{}}Baseline\end{tabular}} & \multirow{2}{*}{Parameters} & \multirow{2}{*}{GFLOPs} & \multicolumn{3}{c}{CMU-Mocap}     \\ \cline{4-6} 
                                                                                &                             &                         & 1sec      & 2sec      & 3sec      \\ \hline
TBIFormer\cite{peng2023trajectory}                                                                       &       6.1M                      &  1.7                     &    8.0       &   13.4        &  19.0         \\
JRT \cite{xu2023joint}                                                                            & 6.7M                               &   1.2                      &    8.3       &   13.9        &     18.5      \\ \hline
                                                                $L$=1      & 4.2M                            & 1.2                        &    8.1       &13.6           &17.9           \\
                                                                           $L$=2     &  4.8M                            & 1.2                        & 8.0           & 13.5           & 17.7          \\
                                                                          $L$=4      &6.2M                             &1.5                         &7.8           &  13.2         & 17.8          \\
                                               $L$=5   & 7.0M
                                                  & 1.8
                                                  &8.1
                                                  &13.5
                                                  &18.2
                                                  \\
                                                  \hline
                                           $L$=3 (Ours)        &   \textbf{5.5M}        & 1.4     &  \textbf{7.8} & \textbf{13.0} & \textbf{17.3} \\ \hline
\end{tabular}
\label{paramater}
\end{table}

\begin{table}[t]
\centering
\caption{Ablation study of hyperedges on CMU-Mocap based on MPJPE. We set up three different strategies to verify our hyperedges. (i) remove short-term hyperedges, denote as "w/o short-term";
(ii) remove long-term hyperedges, denote as "w/o long-term"; (iii) remove spatial hyperedges, denote as "w/o spatial"; (iv) remove all hyperedges and construct a fully-connected graph with $N\times T \times J$ nodes, denote as "fully-connected".}

\begin{tabular}{c|ccc|ccc}
\hline
\multirow{2}{*}{Ablation} & \multicolumn{3}{c|}{CMU-Mocap} & \multicolumn{3}{c}{Mix1} \\ \cline{2-7} 
                          & 1sec     & 2sec     & 3sec     & 1sec   & 2sec   & 3sec   \\ \hline
w/o short-term                 &   8.0       &  13.6        &  18.5        & 13.9       &  24.4      &  30.8      \\
w/o long-term                 &   8.0       &  13.7        &  18.8        & 14.2       &  24.7      &  31.1      \\
w/o spatial                 &  8.1        &   13.8       &    18.9      &  14.2      &  25.0      & 31.8       \\
fully-connected          &  11.2        &  20.6        &  28.7       &   20.5     &   32.8     &  43.9      \\ \hline
Ours                      & \textbf{7.8}         &   \textbf{13.0}        &\textbf{17.3}           &  \textbf{13.6}       & \textbf{23.6}        &
\textbf{29.4} \\ \hline
\end{tabular}

\label{ablation on module}
\end{table}


\subsection{Ablation Study}

The ablation experiments are conducted on the CMU-Mocap and Mix1 datasets, with results presented in Table \ref{paramater}, Table \ref{ablation on module}, and Table \ref{loss}. We analyze the computational complexity and discuss the strategy of hyperedge construction. Additionally, we conduct an ablation study on different loss function settings.
\subsubsection{Computational complexity analysis}
We evaluate the trade-off between the model’s computational
cost and performance, as shown in Table. \ref{paramater}. We
report the number of parameters and an estimate of the floating
operations GFLOPS (Giga Floating-point Operations
Per Second) of the models when predicting 14 frames (900ms)
on the CMU-Mocap dataset. The table shows
that UnityGraph achieves increasingly better
performance from the number of iteration $L$ = 1 to $L$ = 5
and achieves the best results when $L$ = 3. With the further increase of $L$, UnityGraph
obtains inferior performance. Compared to JRT \cite{xu2023joint}, We achieved better performance by using 20\% fewer
parameters.

\subsubsection{Effects of hyperedges} \label{fc}
To demonstrate the effectiveness
of hyperedges, we compare our model with three other settings: i) remove short-term hyperedges;
ii) remove short-term hyperedges; iii) remove long-term hyperedges; iv) remove all hyperedges and construct a fully-connected graph with $N \times T$ nodes. The results are shown in Table. \ref{ablation on module}. It is obvious that the different hyperedges all contribute to the improvement of motion prediction performance. In addition, a fully-connected graph structure can also promote coherence and coupling of spatio-temporal features. However, this structure may compromise accuracy as the number of nodes in the graph increases with the number of individuals ($N$), leading to a graph of large order and size. This increase could make computationally expensive. For a fully connected graph, the total computational complexity for updating all nodes once is $O\left ( \left ( N\times T \right )^{2} \times d\right )$, where $d = 3 \times J$ is the feature dimension of each node. In contrast, for unitygraph, the computational complexity is divided into two parts: updating the hyperedges and updating the nodes. The complexity for updating the three types of hyperedges is given by:
\begin{equation}
     O\left (  N\times T \times d\right ) + O\left (  N\times T \times d\right ) + O\left ( N \times \left (  T-1\right )\times d\right ) 
\end{equation}
The complexity for updating the nodes is
$O\left (  N\times T \times d \times k \right )$,
where $k = 4$ represents the number of hyperedges associated with each node. Thus, the overall computational complexity for the unitygraph is still of the order $O\left (  N\times T \times d\right )$. Despite each node being connected by four hyperedges, the computational effort remains significantly lower than that of a fully connected graph, which is $O\left ( \left ( N\times T \right )^{2} \times d\right )$.
\begin{table}[]
\centering
\caption{Ablation study of loss functions on CMU-Mocap and Mix1 based on MPJPE.}

\begin{tabular}{ccc|ccc|ccc}
\hline
\multicolumn{1}{c|}{\multirow{2}{*}{$\mathcal{L}_{pre}$}} & \multicolumn{1}{c|}{\multirow{2}{*}{$\mathcal{L}_{rec}$}} & \multirow{2}{*}{$\mathcal{L}_{inf}$} & \multicolumn{3}{c|}{CMU-Mocap}    & \multicolumn{3}{c}{Mix1} \\ \cline{4-9} 
\multicolumn{1}{c|}{}                                     & \multicolumn{1}{c|}{}                                     &                                      & 1sec      & 2sec      & 3sec      & 1sec   & 2sec   & 3sec   \\ \hline
$\checkmark$                                              &                                                           &                                      &   8.1        &   13.9        & 18.8          &  13.8      &  24.2      &   31.1     \\
$\checkmark$                                              & $\checkmark$                                              &                                      &  8.0         & 13.6          &  18.0         &   13.8     & 24.1       &  30.8      \\
$\checkmark$                                              &                                                           & $\checkmark$                         &  7.9        &  13.4         &   17.8        &  13.7      & 23.7       & 29.8       \\ \hline
$\checkmark$                                              & $\checkmark$                                              & $\checkmark$                         & \textbf{7.8} & \textbf{13.0} & \textbf{17.3} & \textbf{13.6} & \textbf{23.6} & \textbf{29.4}       \\ \hline
\end{tabular}

\label{loss}
\end{table}

\begin{table}[]
\centering
\caption{Ablation study on weight of loss functions. The best result is achieved when $\mathcal{L}_{pre}=0.70$,$\mathcal{L}_{rec}=0.20$,and $\mathcal{L}_{inf}=0.10$.}
\begin{tabular}{ccc|ccc|ccc}
\hline
\multicolumn{1}{c|}{\multirow{2}{*}{$\mathcal{L}_{pre}$}} & \multicolumn{1}{c|}{\multirow{2}{*}{$\mathcal{L}_{rec}$}} & \multirow{2}{*}{$\mathcal{L}_{inf}$} & \multicolumn{3}{c|}{CMU-Mocap}               & \multicolumn{3}{c}{Mix1}                      \\ \cline{4-9} 
\multicolumn{1}{c|}{}                   & \multicolumn{1}{c|}{}                   &                    & 1sec         & 2sec          & 3sec          & 1sec          & 2sec          & 3sec          \\ \hline
0.10                                     & 0.45                                    & 0.45               & 9.1          & 15.0          & 20.7          & 17.5          & 30.1          & 40.6          \\
0.20                                     & 0.40                                     & 0.40                & 8.7          & 14.5          & 18.9          & 17.2          & 29.6          & 39.8          \\
0.30                                     & 0.30                                    & 0.35               & 8.5          & 14.1          & 18.6          & 16.6          & 27.8          & 35.7          \\
0.40                                    & 0.30                                     & 0.30                & 8.3          & 13.8          & 18.4          & 14.9          & 25.2          & 31.4          \\
0.50                                     & 0.25                                    & 0.25               & 8.1          & 13.6          & 18.1          & 14.2          & 24.4          & 30.7          \\
0.60                                     & 0.20                                     & 0.20                & 7.9          & 13.5          & 17.8          & 13.9          & 24.0          & 30.1          \\
0.60                                     & 0.30                                     & 0.10                & 7.9          & 13.4          & 17.7          & 13.9          & 23.9          & 30.0          \\
0.70                                     & 0.15                                    & 0.15               & 7.8          & 13.2          & 17.6          & 13.7          & 23.9          & 29.8          \\
\textbf{0.70}                            & \textbf{0.20}                            & \textbf{0.10}       & \textbf{7.8} & \textbf{13.0} & \textbf{17.3} & \textbf{13.6} & \textbf{23.6} & \textbf{29.4} \\
0.70                                     & 0.10                                     & 0.20               & 7.8          & 13.2          & 17.5          & 13.7          & 23.8          & 29.7          \\
0.80                                     & 0.10                                     & 0.10                & 7.9          & 13.3          & 17.7          & 13.8          & 23.9          & 29.9          \\
0.90                                     & 0.05                                    & 0.05               & 7.9          & 13.5          & 17.8          & 13.9          & 23.9          & 30.1          \\ \hline
\end{tabular}
\label{weight of loss function}
\end{table}
\subsubsection{Effects of loss funciton}
In this section, we perform extensive
ablation studies on the CMU-Mocap and Mix1 datasets to investigate the
contribution of the different loss functions; see
Table. \ref{loss}. We compare our strategy with three other settings: i) only $\mathcal{L}_{pre}$; ii) $\mathcal{L}_{pre} + \mathcal{L}_{rec}$; iii) $\mathcal{L}_{pre} + \mathcal{L}_{inf}$. We see that each loss function can help UnityGraph get better prediction.

\subsubsection{Weight of loss function}
We conduct the ablation study on weight of loss functions since we chose the weight coefficients manually. We can see that when the value of $\mathcal{L}_{pre}<0.6$ in Table. \ref{weight of loss function}, none of the results perform well. As the weight of
$\mathcal{L}_{pre}$ increases, the best results are achieved at 0.7. This suggests that $\mathcal{L}_{pre}$ plays a dominant role in the prediction, while $\mathcal{L}_{rec}$, $\mathcal{L}_{inf}$ help achieve more accurate predictions.
\section{Conclusion}
In this paper, we introduce a novel graph structure named UnityGraph for multi-person motion prediction. UnityGraph addresses the previous issue of maintaining consistency and coupling of spatio-temporal features, which was caused by separately modeling the spatial and temporal dimensions. Extensive experiments demonstrate that UnityGraph achieves state-of-the-art performance. In contrast to the multi-path modeling strategy adopted by most previous methods,our method offers a novel perspective with a single graph in human motion prediction. In the future, we will fine-tune UnityGraph and conduct experiments on single-person datasets like H3.6M to demonstrate its effectiveness in single-person motion prediction tasks.

\small{
\bibliographystyle{IEEEtran}
\bibliography{egbib}}

\end{document}